\documentclass{ecai}

\usepackage{microtype}
\usepackage{bbm}
\usepackage{graphicx}
\usepackage{mathtools} 
\usepackage{subcaption}
\usepackage{booktabs} 
\usepackage[numbers]{natbib}
\usepackage{hyperref}

\newcommand{\CNF}{\textsf{CNF}}
\newcommand{\DNF}{\textsf{DNF}}
\newcommand{\Rule}{\textsf{Rule}}

\usepackage{amsmath}
\usepackage{amssymb}
\usepackage{mathtools}
\usepackage{amsthm}
\usepackage{hyperref}
\usepackage{url}
\usepackage{amsmath}
\usepackage{amssymb}
\usepackage{mathtools}
\usepackage{amsthm}
\usepackage{graphicx}
\usepackage{multirow}
\usepackage{microtype}
\usepackage{graphicx}
\usepackage{booktabs} 
\usepackage{graphicx}
\usepackage{capt-of}
\usepackage{caption}
\usepackage{booktabs}
\usepackage{varwidth}
\usepackage{svg}            
\usepackage{tikz}
\usepackage{comment}

\usepackage{amsmath,amssymb} 
\usepackage{color}
\usepackage{hyperref}   
\usepackage{booktabs}
\usepackage{multirow}
\usepackage{caption}
\usepackage{svg}  
\usepackage{wrapfig,lscape}
\usepackage{float}
\usepackage{rotating}
\usepackage{multicol}
\usepackage{listings}
\usepackage{tikz}
\usepackage{makecell}
\usepackage{comment}

\usepackage{pgfplots}
\usetikzlibrary{positioning, shapes, arrows, chains, intersections}
\newcommand\xstart{0}

\usetikzlibrary{calc}
\usetikzlibrary{decorations.pathreplacing}
\usetikzlibrary{tikzmark,decorations.pathreplacing,calligraphy, patterns}
\definecolor{jonquil}{rgb}{0.98, 0.85, 0.37}
\definecolor{babypink}{rgb}{0.96, 0.73, 0.73}
\definecolor{babypink2}{rgb}{0.96, 0.82, 0.82}
\definecolor{babyblueeyes}{rgb}{0.57, 0.73, 0.92}
\definecolor{babyblueeyes2}{rgb}{0.63, 0.79, 0.95}
\definecolor{darkseagreen}{rgb}{0.63, 0.83, 0.63}
\definecolor{darkseagreen2}{rgb}{0.63, 0.92, 0.63}

\usepackage[capitalize,noabbrev]{cleveref}

\theoremstyle{plain}

\theoremstyle{definition}

\theoremstyle{remark}

\usepackage{bm}

\usepackage[textsize=tiny]{todonotes}

\newcommand{\myparagraphtitle}[1]{\vspace*{-0.05cm}\paragraph{#1}}
\newlength{\subseclen}
\begin{document}

\begin{frontmatter}

\title{Neural Network-Based Rule Models With Truth Tables}

\author[*]{\fnms{Adrien}~\snm{Benamira}}
\author[*]{\fnms{Tristan}~\snm{Guérand}}
\author{\fnms{Thomas}~\snm{Peyrin}} 
\author{\fnms{Hans}~\snm{Soegeng}}

\address{Nanyang Technological University, Singapore}
\address[*]{Main contribution}

\begin{abstract}
Understanding the decision-making process of a machine/deep learning model is crucial, particularly in security-sensitive applications.  In this study, we introduce a neural network framework that combines the global and exact interpretability properties of rule-based models with the high performance of deep neural networks. 

Our proposed framework, called \textit{Truth Table rules} (TT-rules), is built upon \textit{Truth Table nets} (TTnets), a family of deep neural networks initially developed for formal verification. By extracting the set of necessary and sufficient rules $\mathcal{R}$ from the trained TTnet model (global interpretability), yielding the same output as the TTnet (exact interpretability), TT-rules effectively transforms the neural network into a rule-based model. This rule-based model supports binary classification, multi-label classification, and regression tasks for tabular datasets. Furthermore, our TT-rules framework optimizes the rule set $\mathcal{R}$ into $\mathcal{R}_{opt}$ by reducing the number and size of the rules. To enhance model interpretation, we leverage Reduced Ordered Binary Decision Diagrams (ROBDDs) to visualize these rules effectively.

After outlining the framework, we evaluate the performance of TT-rules on seven tabular datasets from finance, healthcare, and justice domains. We also compare the TT-rules framework to state-of-the-art rule-based methods. Our results demonstrate that TT-rules achieves equal or higher performance compared to other interpretable methods while maintaining a balance between performance and complexity. Notably, TT-rules presents the first accurate rule-based model capable of fitting large tabular datasets, including two real-life DNA datasets with over 20K features. Finally, we extensively investigate a rule-based model derived from TT-rules using the Adult dataset.
\end{abstract}

\end{frontmatter}

\section{Introduction}\label{sec:intro}

Deep Neural Networks (DNNs) have been widely and successfully employed in various machine learning tasks, but concerns regarding their security and trustworthiness persist. One of the primary issues associated with DNNs, as well as ensemble ML models in general, is their lack of explainability and the challenge of incorporating human knowledge into them due to their inherent complexity~\cite{ribeiro2016should, ribeiro2018anchors}. 
Therefore, there is a significant research focus on achieving global and exact interpretability for these systems, especially in safety-critical applications ~\cite{ai2023artificial, eu2021airegulation}. 


In contrast, rule-based models~\cite{freitas2014comprehensible}, including tree-based models~\cite{bessiere2009minimising}, are specifically designed to offer global and exact explanations, providing insights into the decision-making process that yields the same output as the model. However, they generally exhibit lower performance compared to other models like DNNs or ensemble ML model~\cite{kadra2021well}. Additionally, they encounter scalability issues when dealing with large datasets and lack flexibility in addressing various types of tasks, often limited to binary classification~\cite{borisov2022deep}.

To the best of our knowledge, there is currently no family of DNNs that possesses both global and exact interpretability akin to rule-based models, while also demonstrating scalability on real-life datasets without the need for an explainer. This limitation is significant since explainer methods often provide only local, inexact, and potentially misleading explanations~\cite{ribeiro2016should, ribeiro2018anchors, slack2020fooling}.

\paragraph{Our approach.} This paper introduces a novel neural network framework that effectively combines the interpretability of rule-based models with the high performance of DNNs. Our framework, called TT-rules, builds upon the advancements made by Benamira \textit{et al.}~\cite{10.1007/978-3-031-25056-9_31} and Agarwal \textit{et al.}~\cite{agarwal2020neural}. The latter proposed a neural network architecture that achieves interpretability by utilizing several DNNs, each processing a single continuous input feature, and a linear layer for merging them. The effectiveness of aggregating local features on image datasets to achieve high accuracy has been demonstrated by Brendel \textit{et al.}~\cite{brendel2019approximating}. Similarly, Agarwal \textit{et al.}~\cite{agarwal2020neural} showed that aggregating local features on tabular datasets can also yield high accuracy. Furthermore, Benamira \textit{et al.}~\cite{10.1007/978-3-031-25056-9_31} introduced a new Convolutional Neural Network (CNN) filter function called the Learning Truth Table (LTT) block. The LTT block has the unique property of its complete distribution being computable in constant and practical time, regardless of the architecture. This allows the transformation of the LTT block from weights into an exact mathematical Boolean formula. Since an LTT block is equivalent to a CNN filter, the entire neural network model, known as Truth Table Net (TTnet), can itself be represented as a Boolean formula.

To summarize, while Agarwal \textit{et al.}~\cite{agarwal2020neural} focused on continuous inputs, and Benamira \textit{et al.}~\cite{10.1007/978-3-031-25056-9_31} focused on discrete inputs, our approach leverages the strengths of both works to achieve high accuracy while maintaining global and exact interpretability.

\paragraph{Our contributions.} To optimize the rule set $\mathcal{R}$, our TT-rules framework employs two post-training steps. Firstly, we automatically integrate \textit{``Don't Care Terms''} ($DCT$), utilizing human logic, into the truth tables. This reduces the size of each rule in the set $\mathcal{R}$. Secondly, we introduce and analyze an inter-rule correlation score to decrease the number of rules in $\mathcal{R}$. These optimizations, specific to the TT-rules framework, automatically and efficiently transform the set $\mathcal{R}$ into an optimized set $\mathcal{R}{opt}$ in constant time. We also quantify the trade-offs among performance, the number of rules, and their sizes. At this stage, we obtain a rule-based model from the trained DNN TTnet, which can be used for prediction by adding up the rules in $\mathcal{R}{opt}$ according to the binary or floating linear layer. To enhance the interpretability of the model, we convert all rule equations into Reduced Ordered Binary Decision Diagrams.

\paragraph{Our claims.} A) The TT-rules framework demonstrates versatility and effectiveness across various tasks, including binary classification, multi-classification, and regression. A-1) Our experiments encompassed five machine learning datasets: Diabetes~\cite{Dua:2019} in healthcare, Adult~\cite{Dua:2019}, HELOC~\cite{HELOC20XXfico}, and California Housing~\cite{pace1997sparse} in finance, and Compas~\cite{angwin2016machine} in the justice domain. The results clearly indicate that the TT-rules framework surpasses most interpretable models in terms of Area Under Curve/Root Mean Square Error (AUC/RMSE), including linear/logistic regression, decision trees, generalized linearized models, and neural additive models. A-2) On two datasets, the TT-rules framework performs comparably to XGBoost and DNN models. A-3) We conducted a comparative analysis of the performance-complexity tradeoff between our proposed TT-rules framework and other state-of-the-art rule-based models, such as generalized linearized models~\cite{wei2019generalized}, RIPPER~\cite{cohen1995fast,cohen1999simple}, decision trees (DT)\cite{pedregosa2011scikit}, and ORS\cite{yang2021learning}, specifically focusing on binary classification tasks. Our findings demonstrate that the TT-rules framework outperforms all the aforementioned models, except for the generalized linearized models, in terms of the performance-complexity tradeoff.

B) Scalability is a key strength of our model, enabling it to handle large datasets with tens of thousands of features, such as DNA datasets~\cite{Tirosh2016Dissecting, Puram2017SingleCell,liu2018integrated}, which consist of over 20K features. Our model not only scales efficiently but also performs feature reduction, compressing the initial 20K features of the first DNA datasets~\cite{Tirosh2016Dissecting, Puram2017SingleCell} into 1K rules, and reducing the 23K features of the second DNA dataset~\cite{liu2018integrated} into 9K rules.

C) A distinctive feature of our framework lies in its inherent global and exact interpretability. C-1) To showcase its effectiveness, we provide a concrete use case with the Adult dataset and thoroughly investigate its interpretability. C-2) We explore the potential for incorporating human knowledge into our framework. C-3) Additionally, we highlight how experts can leverage the rules to detect concept shifts, further emphasizing the interpretability aspect of our framework.

\paragraph{Outline.} This paper is structured as follows. Section~\ref{sec:Related_work} presents a comprehensive literature review on rule-based models. In Section~\ref{sec:Background}, we establish the notations and fundamental concepts that will be utilized throughout the paper. Section~\ref{sec:TT} offers a detailed analysis of the TT-rules framework, exploring its intricacies and functionalities. In Section~\ref{sec:Results}, we present the experimental results obtained and compare them with the current state-of-the-art approaches. Additionally, we showcase the scalability of our framework and illustrate its applicability through a compelling case study. The limitations of the proposed approach are discussed in Section~\ref{sec:Limitations}, followed by the concluding remarks in Section~\ref{sec:Conclusion}.

\section{Related work}
\label{sec:Related_work}

\subsection{Classical rule-based models}
Rule-based models are widely used for interpretable classification and regression tasks. This class encompasses various models such as decision trees \cite{bessiere2009minimising}, rule lists \cite{rivest1987learning, angelino2017learning, dash2018boolean}, linear models, and rule sets \cite{lakkaraju2016interpretable, cohen1995fast, cohen1999simple, quinlan2014c4, wei2019generalized}. Rule sets, in particular, offer high interpretability due to their straightforward inference process \cite{lakkaraju2016interpretable}. However, traditional rule sets face limitations when applied to large tabular datasets, binary classification tasks, and capturing complex feature relationships. These limitations result in reduced accuracy and limited practicality in real-world scenarios \cite{yang2021learning, wang2021scalable}. To overcome these challenges, we leverage the recent work of Benamira \textit{et al.} \cite{10.1007/978-3-031-25056-9_31}, who proposed an architecture specifically designed to be encoded into $\CNF$ formulas \cite{biere2009handbook}. This approach has demonstrated scalability on large datasets like ImageNet and can be extended to multi-label classification tasks. In this study, our objective is to extend Benamira's approach to handle binary and multi-class classification tasks, as well as regression tasks, across a wide range of tabular datasets ranging from 17 to 20K features. 

\subsection{DNN-based rule models}

There have been limited investigations into the connection between DNNs and rule-based models. Two notable works in this area are DNF-net \cite{abutbul2020dnf} and RRL \cite{wang2021scalable}. DNF-net focuses on the activation function but lacks available code, while RRL specifically addresses classification tasks. Although RRL achieved high accuracy on the Adult dataset, its interpretability raises concerns due to its complex nature, involving millions of terms, and its training process that is time-consuming \cite{wang2021scalable}. Neural Additive Models (NAMs) \cite{agarwal2020neural} represent another type of neural network architecture that combines the flexibility of DNNs with the interpretability of additive models. While NAMs have demonstrated superior performance compared to traditional interpretable models, they do not strictly adhere to the rule-based model paradigm and can pose challenges in interpretation, especially when dealing with a large number of features. In this paper, we conduct a comparative analysis to evaluate the performance and interpretability of our TT-rules framework in comparison to NAMs \cite{agarwal2020neural}. 

\section{Background}
\label{sec:Background}

\subsection{Rule-based models}

\subsubsection{Rules format : $\DNF$ and ROBDD}

Rule-based models are a popular method for generating decision predicates expressed in $\DNF$. For instance, in the Adult dataset \cite{Dua:2019}, a rule for determining whether an individual would earn more than 50K\$/year might look like:
$$((\text{Age}>34) \land \text{Maried}) \lor (\text{Male} \land (\text{Capital Loss} < \text{1k/year})) $$

Although a rule is initially expressed in $\DNF$ format, a decision tree format is often preferred. To achieve this, the $\DNF$ is transformed into its equivalent Reduced Ordered Binary Decision Diagram (ROBDD) graph: a directed acyclic graph used to represent a Boolean function \cite{lee1959representation, akers1978binary, bryant1986graph, aqajari2021pyeda}. 

\subsubsection{Infer a set of rule-based model}

In a binary classification problem, we are presented with a set of rules $\mathcal{R}$ and a corresponding set of weights $\mathcal{W}$. These rules and weights can be separated into two distinct sets, namely $\mathcal{R}{+}$ and $\mathcal{W}{+}$ for class 1, and $\mathcal{R}{-}$ and $\mathcal{W}{-}$ for class 0. Given an input $I$, we can define the rule-based model as follows:

\begin{equation*}
Classifier(I, \mathcal{R}) = \left\{
\begin{array}{ll}
1 & \mbox{if } S_{+}(I) - S_{-}(I) > 0 \\
0 & \mbox{otherwise.}
\end{array}
\right.
\end{equation*}

Here, $S_{+}(I)$ and $S_{-}(I)$ denote the scores for class 1 and class 0, respectively. These scores are calculated using the following equations:

\begin{equation*}
\left\{
\begin{array}{ll}
S_{+}(I) = \sum_{(r_{+}, w_{+}) \in (\mathcal{R}{+}, \mathcal{W}{+})} w_{+} \times \mathbbm{1}_{ { r{+}(I) \texttt{ is True} } } \\
S_{-}(I) = \sum_{(r_{-}, w_{-}) \in (\mathcal{R}{-}, \mathcal{W}{-})} w_{-} \times \mathbbm{1}_{ { r{-}(I) \texttt{ is True} } }
\end{array}
\right.
\end{equation*}

\noindent where $\mathbbm{1}_{ { r(I) \texttt{ is True} } }$ represents the binary indicator that is equal to 1 if the input $I$ satisfies the rule $r$, and 0 otherwise. This rule-based model can be easily extended to multi-class classification and regression tasks. 

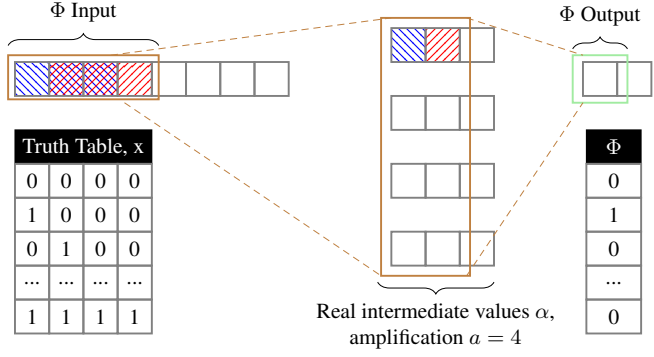
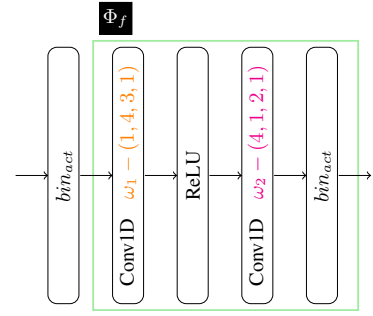
\begin{figure*}[htb!]
  \centering
\begin{subfigure}[t]{0.55\textwidth}
\centering
\scalebox{0.9}{
\begin{tikzpicture}

\foreach \x in {0,...,0}
  \draw[gray, thick,pattern=north west lines, pattern color=blue] (0 + \x*0.5,0.5) rectangle (0.5 + \x*0.5,1);

\foreach \x in {1,...,2}
  \draw[gray, thick, pattern=north east lines, pattern color=red, preaction={pattern=north west lines, pattern color=blue}] (0 + \x*0.5,0.5) rectangle (0.5 + \x*0.5,1);

\foreach \x in {3,...,3}
  \draw[gray, thick,pattern=north east lines, pattern color=red] (0 + \x*0.5,0.5) rectangle (0.5 + \x*0.5,1);

\foreach \x in {4,...,7}
  \draw[gray, thick] (0 + \x*0.5,0.5) rectangle (0.5 + \x*0.5,1);

    \draw[gray, thick,pattern=north west lines, pattern color=blue] (5.5 + 0*0.5,1 - 1*0) rectangle (6+ 0*0.5,1.5- 1*0);
    \draw[gray, thick,pattern=north east lines, pattern color=red] (5.5 + 1*0.5,1 - 0) rectangle (6+ 1*0.5,1.5- 0);
    \draw[gray, thick] (5.5 + 2*0.5,1 - 0) rectangle (6+ 2*0.5,1.5- 0);

\foreach \y in {1,...,3}
  \foreach \x in {0,...,2}
    \draw[gray, thick] (5.5 + \x*0.5,1 - \y) rectangle (6+ \x*0.5,1.5- \y);

\foreach \x in {0,1}
  \draw[gray,thick] (8.5+\x*0.5-0.2,0.9-0.4) rectangle (9+\x*0.5-0.2,1.4-0.4); 

\draw [decorate,decoration={brace,amplitude=5pt,raise=1ex}]
 (7,-2.1) -- (5-0.2 + 1*0.5,-2.1) node[midway,yshift= -2.5em, align=center]{Real intermediate values $\alpha$,\\amplification $a=4$};

\draw[brown, thick] (5.35 + 0*0.5,1-3.15) rectangle (6.65,1.65);

\draw[darkseagreen2, thick] (8.35+0*0.5-0.2,1.5-0.4) rectangle (9.15+0*0.5-0.2,0.8-0.4);
\draw[brown, densely dashed] (5.5 + 1*0.5+0.6,1.65+0.0) -- (8.5+0*0.5-0.2,0.6+0.9-0.4);
\draw[brown, densely dashed] (5.5 + 1*0.5+0.6,1-3.15) -- (8.5+0*0.5-0.2,-0.1+0.9-0.4);
\draw[brown, densely dashed] (-0.1+ 3*0.5,0.6+0.5) -- (5.5,1.65);
\draw[brown, densely dashed] (0.1 + 3*0.5,-0.1+0.5) -- (5.35+0*0.5,1-3.15-0);
\draw[brown, thick] (-0.1,0.65-0.2) rectangle (0.1 + 4*0.5,1.1);

\draw [decorate,decoration={brace,amplitude=5pt,raise=1ex}]
 (-0.1,1.1) -- (0.1 + 4*0.5,1.1) node[midway,yshift= 2em]{$\Phi$ Input};

\draw [decorate,decoration={brace,amplitude=5pt,raise=1ex}]
 (-0.15+8.5-0.2,0.9+1-0.5-0.4+0.05) -- (0.15 + 1*0.5+8.5-0.2,0.9+1-0.5-0.4+0.05) node[midway,yshift= 2em]{$\Phi$ Output};
 


\draw[black, fill=black] (0, -0.5) rectangle (0 + 4*0.5,0) node[pos=0.5,text=white] {\small Truth Table, x};

\foreach \x in {0,...,3}
  \draw[gray, thick] (0 + \x*0.5,-0.5-1*0.5) rectangle (0.5 + \x*0.5,-0.-1*0.5) node[pos = 0.5, black] {0};
\foreach \x in {1,...,3}  
  \draw[gray, thick] (0 + \x*0.5,-0.5-2*0.5) rectangle (0.5 + \x*0.5,-0.-2*0.5) node[pos = 0.5, black] {0};
\draw[gray, thick] (0 + 0*0.5,-0.5-2*0.5) rectangle (0.5 + 0*0.5,-0.-2*0.5) node[pos = 0.5, black] {1};  
\draw[gray, thick] (0 + 0*0.5,-0.5-3*0.5) rectangle (0.5 + 0*0.5,-0.-3*0.5) node[pos = 0.5, black] {0};
\foreach \x in {2,...,3}  
  \draw[gray, thick] (0 + \x*0.5,-0.5-3*0.5) rectangle (0.5 + \x*0.5,-0.-3*0.5) node[pos = 0.5, black] {0};
\draw[gray, thick] (0 + 1*0.5,-0.5-3*0.5) rectangle (0.5 + 1*0.5,-0.-3*0.5) node[pos = 0.5, black] {1};  
\foreach \y in {4}
\foreach \x in {0,...,3}
  \draw[gray, thick] (0 + \x*0.5,-0.5-\y*0.5) rectangle (0.5 + \x*0.5,-0.-\y*0.5) node[pos = 0.5, black] {...};
\foreach \y in {5}
  \foreach \x in {0,...,3}
    \draw[gray, thick] (0 + \x*0.5,-0.5-\y*0.5) rectangle (0.5 + \x*0.5,-0.-\y*0.5) node[pos = 0.5, black] {1};

\draw[black, fill=black] (8.35, -0.5) rectangle (9.15,0.0) node[pos=0.5,text=white] {$\Phi$};
\draw[gray, thick] (8.35,-0.5-1*0.5) rectangle (9.15,-0-1*0.5) node[pos = 0.5, black] {0};
\draw[gray, thick] (8.35,-0.5-2*0.5) rectangle (9.15,-0.-2*0.5) node[pos = 0.5, black] {1};
\draw[gray, thick] (8.35,-0.5-3*0.5) rectangle (9.15,-0.-3*0.5) node[pos = 0.5, black] {0};
\draw[gray, thick] (8.35,-0.5-4*0.5) rectangle (9.15,-0.-4*0.5) node[pos = 0.5, black] {...};
\draw[gray, thick] (8.35,-0.5-5*0.5) rectangle (9.15,-0.-5*0.5) node[pos = 0.5, black] {0};
\end{tikzpicture}
}
  \caption{LTT filter $\Phi_f$ computation and transformation into a truth table. $\Phi_f$ is characterized by weights $\textcolor{orange}{\omega_1}$ with parameters (input channel, output channel, kernel size, stride) = $(1,4,3,1)$, and $\textcolor{magenta}{\omega_2} = (4,1,2,1)$. $\Phi_f$ input bit-size is $4$ (i.e., \textcolor{brown}{brown} box) since the output feature (i.e., \textcolor{darkseagreen2}{green} box) requires $4$ input entries (i.e., \textcolor{blue}{blue} \& \textcolor{red}{red} hashed boxed). 
  }
  \label{fig: Amplification Layer}
\end{subfigure}
\hspace{0.4cm}
\begin{subfigure}[t]{0.35\textwidth}
\centering
\scalebox{0.85}{
\begin{tikzpicture}

\draw[black, fill=black] (-1.2+2, 4.7-0.5) rectangle (-1.2+0.5+2,4.7+0.0) node[pos=0.50,text=white] {$\Phi_f$};

\draw[darkseagreen2, thick] (-0.3+1,0-0.1) rectangle (5-0.2,4.1);

\foreach \x in {0,...,4} {
  \draw[rounded corners] (0+ \x, 0) rectangle (0.48+\x, 4);
  }
\foreach \x in {0,...,5} {
  \draw[->] (0.5+ \x -1, 2) -- (1+\x-1, 2);
  }
\path (0, 0) -- node[rotate=90,anchor=center] {$bin_{act}$} (0.5, 4);
\path (0+1, 0) -- node[rotate=90,anchor=center] {Conv1D \:\: $\textcolor{orange}{\omega_1 - (1,4,3,1)}$ }(0.5 + 1, 4);
\path (0+2, 0) -- node[rotate=90,anchor=center] {ReLU}(0.5+2, 4);
\path (0+3, 0) -- node[rotate=90,anchor=center] {Conv1D \:\: $\textcolor{magenta}{\omega_2 - (4,1,2,1)}$}(0.5+3, 4);
\path (0+4, 0) -- node[rotate=90,anchor=center] {$bin_{act}$}(0.5+4, 4);
\end{tikzpicture}
}
  \caption{LTT overview in one dimension. The intermediate values and the weights ($\textcolor{orange}{\omega_1}$, $\textcolor{magenta}{\omega_2}$) are real and the input/output values are binary. $bin_{act}$: Heaviside step function defined as $bin_{act}(x) = \frac{1}{2} + \frac{sgn(x)}{2}$ with $x \in \mathbb{R}$. }
  \label{fig:LTTAutoEncoder}
\end{subfigure}
\caption{A Learning Truth Table (LTT) filter example in one dimension. 
}

\label{fig: inner and outer Learning Truth Table Block}
\end{figure*}

\subsubsection{Comparing rule-based models}
\label{subsubsec:complexity}

When comparing rule-based models, it is common to evaluate their quality based on three main criteria. The first is their performance, which can be measured using metrics such as AUC, accuracy, or RMSE. The second criterion is the number of rules used in the model. Finally, the overall complexity of the model is also taken into account, which is given as the sum of the size of each rule, for all rules in the model~\cite{freitas2014comprehensible}.

\subsection{Truth Table net (TTnet)}

\begin{figure}[h]
\centering
\captionof{table}{\label{table:TT_example} Truth Table of the LTT block $\Phi_f$ characterized by the weights $\textcolor{orange}{W_1}$ and $\textcolor{magenta}{W_2}$ with $L=5$ and binary input feature names [Is the Sex Male? (Male), Did the person go to University? (Go Uni.), Is the person married? (Married), Is the person born in the US? (Born US), Is the person born in the UK? (Born UK)].}
\begin{minipage}{0.1\textwidth}
\centering
\begin{tabular}{|c|c|c|c||c|}
\hline
$x_0$ & $x_1$ & $x_2$ & $x_3$ & $\Phi_f$ \\
\hline
0 & 0 & 0 & 0 & 0 \\
0 & 0 & 0 & 1 & 1 \\
0 & 0 & 1 & 0 & 0 \\
0 & 0 & 1 & 1 & 0 \\
0 & 1 & 0 & 0 & 0 \\
0 & 1 & 0 & 1 & 0 \\
0 & 1 & 1 & 0 & 0 \\
0 & 1 & 1 & 1 & 0 \\
1 & 0 & 0 & 0 & 0 \\
1 & 0 & 0 & 1 & 0 \\
1 & 0 & 1 & 0 & 0 \\
1 & 0 & 1 & 1 & 0 \\
1 & 1 & 0 & 0 & 0 \\
1 & 1 & 0 & 1 & 0 \\
1 & 1 & 1 & 0 & 0 \\
1 & 1 & 1 & 1 & 0 \\
\hline
\end{tabular}
\end{minipage}%
\hfill%
\begin{minipage}{0.25\textwidth}
\centering
\scalebox{1}{
\begin{tabular}{|c|}
\hline
$\textcolor{orange}{W_1}=$
$\begin{pmatrix}
10 & -1 & 3 \\
6 & -5 & 4 \\
4 & 4 & -3 \\
4 & 4 & 3
\end{pmatrix}$ \\
$\textcolor{magenta}{W_2}=$
$\begin{pmatrix}
-5 & 0 & 9 & -5 \\
-5 & 4 & 0 & 0 \\
\end{pmatrix}$ \\
\hline
\end{tabular}
}

\medskip

\scalebox{0.85}{
\begin{tabular}{|c|}
\hline
$\Phi_f$ expression in \DNF  \\
$x_3  \wedge \overline{x_0} \wedge \overline{x_1} \wedge \overline{x_2}$
\\
\hline
Input space  \\
\resizebox{\hsize}{!}{[Male, Go Uni., Married, Born US, Born UK]} \\
\hline
$\Rule_{0}^{\DNF}$ \\
\resizebox{\hsize}{!}{
$\text{Born US}  \wedge\overline{\text{Male}} \wedge \overline{\text{Go Uni.} } \wedge \overline{\text{Married}}$} \\
\hline
$\Rule_{1}^{\DNF}$ \\
\resizebox{\hsize}{!}{
$\text{Born UK}  \wedge\overline{\text{Go Uni.}} \wedge \overline{\text{Married} } \wedge \overline{\text{Born US}}$} \\
\hline
$\Rule_{1, reduced}^{\DNF}$ \\
\resizebox{\hsize}{!}{
$\text{Born UK}  \wedge\overline{\text{Go Uni.}} \wedge \overline{\text{Married} }$} \\
\hline
\end{tabular}
}
\end{minipage}
\end{figure}

The paper~\cite{10.1007/978-3-031-25056-9_31} proposed a new CNN filter function called Learning Truth Table (LTT) block for which one can compute the complete distribution in practical and constant time, regardless of the network architecture. Then, this LTT block is inserted inside a DNN as CNN filters are integrated into deep convolutional neural networks. 

\subsubsection{Overall LTT design}

An LTT block must meet two essential criteria:

\begin{itemize}
\item (A) The LTT block distribution must be entirely computable in practical and constant time, regardless of the complexity of the DNN.
\item (B) Once LTT blocks are assembled into a layer and layers into a DNN, the latter DNN should be scalable, especially on large-scale datasets such as ImageNet.
\end{itemize}

To meet these criteria, Benamira \textit{et al.}~\cite{10.1007/978-3-031-25056-9_31} proposed the following LTT design rules:

\begin{enumerate}
\item Reduce the input size of the CNN filter to $n \leq 9$.
\item Use binary inputs and outputs.
\item Ensure that the LTT block function uses a nonlinear function.
\end{enumerate}

As a result, each filter in our architecture becomes a truth table with a maximum input size of 9 bits.

\paragraph{Notations.}

We denote the $f^{th}$ 1D-LTT of a layer with input size $ n$, stride $s$, and no padding as $\Phi_f$. Let the input feature with a single input channel $chn_{input}=1$ be represented as $(v_0 \dots v_{L-1})$, where $L$ is the length of the input feature. We define $y_{i, f}$ as the output of the function $\Phi_f$ at position $i$:  

$$y_{i, f} = \Phi_f(v_{i \times s}, v_{i \times s+1}, \dots, v_{i \times s + (n-1)})$$ 

Following the aforementioned rules (1) and (2), $y_{i, f}$ and $(v_{i \times s}, v_{i \times s+1}, \dots, v_{i \times s + (n-1)})$ are binary values, and $n \leq 9$. As a result, we can express the 1D-LTT function $\Phi_f$ as a truth table by enumerating all $2^n$ possible input combinations. The truth table can then be converted into an optimal (in terms of literals) $\DNF$ formula using the Quine–McCluskey algorithm \cite{blake1938} for interpretation.

\begin{figure*}[!htb]
\center
\captionof{table}{\label{tab:perfs} Comparison machine learning dataset of our method to Linear/Logistic Regression)~\cite{pedregosa2011scikit}, Decision Trees (DT)~\cite{pedregosa2011scikit}, GL~\cite{wei2019generalized}, NAM~\cite{agarwal2020neural}, XGBoost~\cite{Chen_2016} and DNNs. Results are obtained with a large TT-rules model, without optiizations. Means and standard deviations are reported from 5-fold cross validation.}
\resizebox{1.9\columnwidth}{!}{
\begin{tabular}{@{}l|c|ccc|c@{}}
\toprule
 & \textbf{Regression} (RMSE)  & \multicolumn{3}{c|}{\textbf{Binary classification} (AUC)}  & \textbf{Multi-classification} (Accuracy) \\ \midrule
 & California Housing & Compas         & Adult          & HELOC          & Diabetes                                             \\ 
 continous/binary \# & 8/144 features & 9/17 features         & 14/100 features          & 24/330 features          & 43/296 features                                             \\ \midrule
Linear/ log          & 0.728 $\pm$ 0.015     & 0.721 $\pm$ 0.010  & 0.883 $\pm$ 0.002 & 0.798 $\pm$ 0.013 & 0.581 $\pm$ 0.002                                                  \\
DT       & 0.514 $\pm$ 0.017     & 0.731 $\pm$ 0.020  & 0.872 $\pm$ 0.002 & 0.771 $\pm$ 0.012 & 0.572 $\pm$ 0.002                                                  \\
GL                   & 0.425 $\pm$ 0.015     & 0.735 $\pm$ 0.013 & 0.904 $\pm$ 0.001 & 0.803 $\pm$ 0.001  & NA                                                  \\
NAM                  & 0.562 $\pm$ 0.007     & 0.739 $\pm$ 0.010 & -              & -              & -                                            \\
TT-rules (Ours)                 & 0.394 $\pm$ 0.017     & 0.742 $\pm$ 0.007 & 0.906 $\pm$ 0.005 & 0.800 $\pm$ 0.001            & 0.584 $\pm$ 0.003                                       \\ \hline 
XGBoost              & 0.532 $\pm$ 0.014     & 0.736 $\pm$ 0.001  & 0.913 $\pm$ 0.002  & 0.802 $\pm$ 0.001  & 0.591 $\pm$ 0.001                                                  \\
DNNs                 & 0.492 $\pm$ 0.009     & 0.732 $\pm$ 0.004  & 0.902 $\pm$ 0.002  & 0.800 $\pm$ 0.010  & 0.603 $\pm$ 0.004                                                   \\ \bottomrule
\end{tabular}}
\end{figure*}

\paragraph{Example 1: From LTT weights to truth table and $\DNF$.}

In this example, we consider a pre-trained 1D-LTT $\Phi_f$ with input size $n=4$, a stride of size $1$, and no padding. The architecture of $\Phi_f$ is given in Figure~\ref{fig:LTTAutoEncoder} composed of two CNN filter layers: the first one has parameters $\textcolor{orange}{W_1}$ with (input channel, output channel, kernel size, stride) = $(1,4,3,1)$, while the second $\textcolor{magenta}{W_2}$ with $(4,1,2,1)$. The inputs and outputs of $\Phi_f$ are binary, and we denote the inputs as [$x_0$, $x_1$, $x_2$, $x_3$]. To compute the complete distribution of $\Phi_f$, we generate all $2^4=16$ possible input/output pairs, as shown in Figure~\ref{fig: Amplification Layer}, and obtain the truth table in Table~\ref{table:TT_example}. This truth table fully characterizes the behavior of $\Phi_f$. We then transform the truth table into a $\DNF$ using the Quine-McCluskey algorithm~\cite{blake1938}. This algorithm provides an optimal (in terms of literals) $\DNF$ formula that represents the truth table. The resulting $\DNF$ formula for $\Phi_f$ can be used to compute the output of $\Phi_f$ for any input. Overall, this example demonstrates the applicability of LTT design rules in the construction of DNNs, as it meets both criteria of LTT blocks being computable in constant time and DNN scalability on large datasets.

 \subsubsection{Overall TTnet design}

 We integrated LTT blocks into the neural network, just as CNN filters are integrated into a deep convolutional neural network: each LTT layer is composed of multiple LTT blocks and there are multiple LTT layers in total. Additionally, there is a pre-processing layer and a final layer. These two layers provide flexibility in adapting to different applications: scalability, formal verification, and logic circuit design.

\section{Truth Table Rules (TT-rules)}\label{sec:TT}

The Truth Table rules framework consists of three essential components. The first step involves extracting the precise set of rules $\mathcal{R}$ once the TTnet has been trained. Next, we optimize $\mathcal{R}$ by reducing the rule's size through \textit{Don't Care Terms} (DCT) injection. At this point, $\mathcal{R}$ is equivalent to the Neural Netwok model: inferring with $\mathcal{R}$ is the same as inferring with the model. Last, we minimize the number of rules using the Truth Table correlation metric. Both techniques serve to enhance the model's complexity while minimizing any potential loss of accuracy.

\subsection{From LTT block to set of rules $\mathcal{R}$} 
\label{subsubsec:setofrule}
\paragraph{General.} We now introduce a method to convert $\Phi_f$ from the general $\DNF$ form into rule set $\mathcal{R}$. In the previous section, we described the general procedure for transforming an LTT block into a $\DNF$ logic gate expression. This expression is independent of the spatial position of the feature. This means that we have:

\begin{equation*}
\left\{
\begin{array}{ll}
y_{0, f} = \Phi_f(v_{0}, v_{1}, \dots, v_{ (n-1)})) \\
... \\
y_{i, f} = \Phi_f(v_{i \times s}, v_{i \times s+1}, \dots, v_{i \times s + (n-1)})) \\
... \\
y_{ \lfloor \frac{L-n}{s} \rfloor, f}  = \Phi_f(v_{L-n}, v_{L-n+1}, \dots, v_{L-1}))
\end{array}
\right.
\end{equation*}

When we apply the LTT $\DNF$ expression to a specific spatial position on the input, we convert the $\DNF$ into a rule. To convert the general $\DNF$ form into a set of rules $\mathcal{R}$, we divide the input into patches and substitute the $\DNF$ literals with the corresponding feature names. The number of rules for one filter corresponds to the number of patches: $\lfloor \frac{L-n}{s} \rfloor$. An example of this process is given in Table~\ref{table:TT_example} and one is provided below.

\paragraph{Example 2: conversion of $\DNF$ expressions to rules.} We established the $\Phi_f$ expression in $\DNF$ form as $ x_3  \wedge \overline{x_0} \wedge \overline{x_1} \wedge \overline{x_2}$. To obtain the rules, we need to consider the padding and the stride of the LTT block. Consider the following 5-feature binary input ($L=5$): [Male, Go Uni., Married, Born in US, Born in UK]. In our case, with a stride at 1 and no padding, we get 2 patches: [Male, Go Uni., Married, Born US] and [Go Uni., Married, Born US, Born UK]. After the substitution of the literal by the corresponding feature, we get 2 rules $\mathcal{R} = \{ \Rule_{0}^{\DNF}, \Rule_{1}^{\DNF} \}$: 

\begin{equation*}
\left\{
\begin{array}{ll}
\Rule_{0, f}^{\DNF} = \text{Born US}  \wedge \overline{\text{Male}} \wedge \overline{\text{Go Uni.} } \wedge \overline{\text{Married}} \\
\Rule_{1, f}^{\DNF} = \text{Born UK}  \wedge\overline{\text{Go Uni.}} \wedge \overline{\text{Married} } \wedge \overline{\text{Born US}} \\
\end{array}
\right.
\end{equation*}

and therefore, the output of the LTT block $\Phi_f$ becomes:

\begin{equation*}
\left\{
\begin{array}{ll}
y_{0, f} = \Rule_{0, f}^{\DNF}(v_{0}, v_{1}, v_{2}, v_{3}) \\
 y_{1, f} = \Rule_{1, f}^{\DNF}(v_{1}, v_{2}, v_{3}, v_{4}) \\
\end{array}
\right.
\end{equation*}

We underline the logic redundancy in $\Rule_{1}^{\DNF}$: if someone is born in the UK, he/she is necessarily not born in the US. We solve this issue by injecting \textit{Don't Care Terms} ($DCT$) into the truth table as we will see in the next section.

\subsection{Automatic post-training optimizations: from $\mathcal{R}$ to $\mathcal{R}_{opt}$}
\label{subsubsec:Optimizations}

In this subsection, we present automatic post-training optimizations that are unique to our model and require the complete computation of the LTT truth table.

\subsubsection{Reducing the rule's size with \textit{Don't Care Terms} ($DCT$) injection} We propose a method for reducing the size of rules by injecting \textit{Don't Care Terms} ($DCT$) into the truth table. These terms represent situations where the LTT block output can be either 0 or 1 for a specific input, without affecting the overall performance of the DNN. We use the Quine-McCluskey algorithm to assign the optimal value to the $DCT$ and reduce the $\DNF$ equations. These $DCT$ can be incorporated into the model either with background knowledge or automatically with the one hot encodings and the Dual Step Function described in the TTnet paper~\cite{10.1007/978-3-031-25056-9_31}.

To illustrate this method, we use Example 2 where we apply human common sense and reasoning to inject $DCT$ into the truth table. For instance, since no one can be born in both the UK and the US at the same time, the literals $x_2$ and $x_3$ must not be 1 at the same time for the second rule. By injecting $DCT$ into the truth table as $[0, 0, 1, DCT, 0, 0, 0, DCT, 0, 0, 0, DCT, 0, 0, 0, DCT]$, we obtain the new reduced rule: $\Rule_{1, reduced}^{\DNF} = \text{Born UK} \wedge\overline{\text{Go Uni.}} \wedge \overline{\text{Married}}$. This method significantly decreases the size of the rules while maintaining the same accuracy, as demonstrated in Table~\ref{tab:res_tab} in Section~\ref{sec:Results}.

\subsubsection{Reducing the number of rules with Truth Table Correlation metric} 

To reduce the number of rules obtained with the TT-rules framework, we introduce a new metric called Truth Table Correlation ($TTC$). This metric addresses the issue of rule redundancy by measuring the correlation between two different LTT blocks, which may learn similar rules since they are completely decoupled from each other. The idea is to identify and remove redundant rules and keep only the most relevant ones.

The $TTC$ metric is defined as follows:

\begin{equation*}
\resizebox{\hsize}{!}{$
TTC(y_1, y_2) = \left\{
    \begin{array}{ll}
        \frac{HW(y_1, \overline{y_2})}{|y_1|} - 1 & \mbox{if } abs(\frac{HW(y_1, \overline{y_2})}{|y_1|} - 1) > \frac{HW(y_1, \overline{y_2})}{|y_1|} \\
        \frac{HW(y_1, y_2)}{|y_1|} & \mbox{otherwise.}
    \end{array}
\right.
$}
\end{equation*}

Here, $y_1$ and $y_2$ are the outputs of the LTT blocks, $\overline{y_2}$ is the negation of $y_2$, $|y_1|$ represents the number of elements in $y_1$, and $HW$ is the Hamming distance function. The Hamming distance between two equal-length strings of symbols is the number of positions at which the corresponding symbols are not equal. The $TTC$ metric varies from -1 to 1. When $TTC=-1$, the LTT blocks are exactly opposite, while they are the same if $TTC=1$. We systematically filter redundant rules with a threshold correlation of $ \pm 0.9$. If the correlation is positive, we delete one of the two filters and give the same value to the second filter. If the correlation is negative, we delete one of the two filters and give the opposite value to the second filter. By using this metric, we can reduce the number of rules and optimize the complexity of the model while minimizing accuracy degradation.

\subsection{Overall TT-rules architecture}

\subsubsection{Pre-processing and final layer}
To maintain interpretability, we apply batch normalization and a step function layer consisting of a single linear layer. The batch normalization allows to learn the thresholds for the continuous features (such as the condition $\text{YoE} > 11$ in Fig.~\ref{fig:casestudy}). We propose two types of training for the final linear layer. The first uses a final sparse binary layer, which forces all weights to be binary and sparse according to a BinMask as in \cite{jia2020efficient}. In order to train without much loss in performance when using the Heaviside step function, Benamira \textit{et al.}~\cite{10.1007/978-3-031-25056-9_31} adopted the Straight-Through Estimator (STE) proposed by~\cite{hubara2016binarized}. The second is designed for scalability and employs floating-point weights, which allows to extend the model to regression tasks. To reduce overfitting, a dropout function is applied in the second case.

\subsubsection{Estimating complexity before training}
In our TT-rules framework, the user is unable to train a final rule-based model with a fixed and pre-selected complexity. However, the complexity can be estimated. The number of rules is determined by multiplying the number of filters $F$ by the number of patches $\lfloor \frac{L-n}{s} \rfloor$. The complexity of each rule is based on the size of the function $n$, and on average, we can expect $n 2^{n-1}$ Boolean gates per rule, before $DCT$ injection. Therefore, the overall complexity is given by $n  \times 2^{n-1} \times \lfloor \frac{L-n}{s} \rfloor \times F$.

\subsubsection{Training and extraction time}
\paragraph{Training.} Compared to other rule-based models, our architecture scales well in terms of training time. The machine learning tabular dataset can be trained in 1-5 minutes for 5-fold cross-validation. For large DNA tabular datasets, our model can be trained in 45 minutes for 5-fold cross-validation, which is not possible with other rule-based models such as GL and RIPPER.

\paragraph{Extraction time for $\mathcal{R}_{opt}$.} Our model is capable of extracting optimized rules at a fast pace. Each truth table can be computed in $2^n$ operations, where $n \leq 9$ is in terms of complexity. In terms of time, our model takes 7 to 17 seconds for Adult~\cite{Dua:2019}, 7 to 22 seconds for Compas~\cite{angwin2016machine}, and 20 to 70 seconds for Diabetes~\cite{Dua:2019}.

\section{Results}\label{sec:Results}

In this section, we present the results of applying the TT-rules framework to seven datasets, which allow us to demonstrate the effectiveness of our approach and provide evidence for the three claims stated in the introduction.

\begin{figure*}[!htb]
\captionof{table}{\label{tab:acc_vs_complexity} Accuracy and complexity on the Compas, Adult and HELOC datasets for different methods. All the TT-rules are computed with our automatic post-training optimizations as described in Section~\ref{subsubsec:Optimizations}. TT-rules big refers to a TTnet trained with a final linear regression with weights as floating points, whereas TT-rules small refers to a TTnet trained with a sparse binary linear regression.}
\center

\resizebox{1.95\columnwidth}{!}{

\begin{tabular}{@{}l|ccc|ccc|ccc@{}}
\toprule
& \multicolumn{3}{c|}{\textbf{Compas}} & \multicolumn{3}{c|}{\textbf{Adult}} & \multicolumn{3}{c}{\textbf{HELOC}} \\ \midrule
& Accuracy & Rules & Complexity & Accuracy & Rules & Complexity & Accuracy & Rules & Complexity \\ \midrule
GL & 0.685 $\pm$ 0.012 & 16 $\pm$ 2 & 20 $\pm$ 6 & 0.852 $\pm$ 0.001 & 16 $\pm$ 1 & 23 $\pm$ 1 & 0.732 $\pm$ 0.001 & 104 $\pm$ 5 & 104 $\pm$ 5 \\
RIPPER & 0.560 $\pm$ 0.006 & 12 $\pm$ 2 & 576 $\pm$ 48 & 0.833 $\pm$ 0.009 & 43 $\pm$ 15 & 14154 $\pm$ 4937 & 0.691 $\pm$ 0.019 & 17 $\pm$ 4 & 792 $\pm$ 186 \\
DT & 0.673 $\pm$ 0.015 & 78 $\pm$ 1 & 12090 $\pm$ 155 & 0.837 $\pm$ 0.004 & 398 $\pm$ 5 & 316410 $\pm$ 3975 & 0.709 $\pm$ 0.011 & 70 $\pm$ 1 & 9522 $\pm$ 136 \\
ORS & 0.670 $\pm$ 0.015 & 11 $\pm$ 1 & 460 $\pm$ 42 & 0.844 $\pm$ 0.006 & 9 $\pm$ 3 & 747 $\pm$ 249 & 0.704 $\pm$ 0.012 & 16 $\pm$ 6 & 1888 $\pm$ 708 \\ \hline
TT-rules big (Ours) & 0.687 $\pm$ 0.005 & 42 $\pm$ 3 & 4893 $\pm$ 350 & 0.851 $\pm$ 0.003 & 288 $\pm$ 12 & 22896 $\pm$ 954 & 0.733 $\pm$ 0.010 & 807 $\pm$ 30 & 103763 $\pm$ 3857 \\
TT-rules small (Ours) & 0.664 $\pm$ 0.013 & 13 $\pm$ 2 & 155 $\pm$ 22 & 0.842 $\pm$ 0.003 & 130 $\pm$ 10 & 673 $\pm$ 145 & 0.727 $\pm$ 0.010 & 82 $\pm$ 30 & 574 $\pm$ 210 \\ \bottomrule
\end{tabular}

}
\end{figure*}

\subsection{Experimental set-up}
\label{subsec:expesetup}

\paragraph{Evaluation measures and training conditions.} We used RMSE, AUC, and accuracy for the evaluation of the regression, binary classification, and multi-class classification respectively. Rules and complexity are defined in Section~\ref{subsubsec:complexity}. All results are presented after grid search and 5-fold cross-validation. All the training features are detailed in the supplementary Section. We compare the performance of our method with that of several other algorithms, including Linear/Logistic Regression~\cite{pedregosa2011scikit}, Decision Trees (DT)\cite{pedregosa2011scikit}, Generalized Linear Models (GL)\cite{wei2019generalized}, Neural Additive Models (NAM)\cite{agarwal2020neural}, XGBoost\cite{Chen_2016}, and Deep Neural Networks (DNNs)~\cite{pedregosa2011scikit}. The supplementary materials provide details on the training conditions used for these competing methods.
Experiments are available on demand.
Our workstation consists of eight cores Intel(R) Core(TM) i7-8650U CPU clocked at 1.90 GHz, 16 GB RAM. 

\vspace*{-0.4cm}

\paragraph{Machine learning datasets.} We utilized a variety of healthcare and non-healthcare datasets for our study. For multi-classification, we used the Diabetes 130 US-Hospitals dataset\footnote{\href{https://bit.ly/diabetes_130_uci}{https://bit.ly/diabetes\_130\_uci}} from the UCI Machine Learning Repository \cite{Dua:2019}. For binary classification tasks, we used two single-cell RNA-seq analysis datasets, one for head and neck cancer\footnote{\href{https://bit.ly/neck_head_rna}{https://bit.ly/neck\_head\_rna}} \cite{Puram2017SingleCell} and another for melanoma \footnote{\href{https://bit.ly/melanoma_rna}{https://bit.ly/melanoma\_rna}}\cite{Tirosh2016Dissecting}, as well as the TCGA lung cancer dataset\footnote{\href{https://bit.ly/tcga_lung_rna}{https://bit.ly/tcga\_lung\_rna}} \cite{liu2018integrated} for regression. For binary classification tasks, we used the Adult dataset\footnote{\href{https://bit.ly/Adult_uci}{https://bit.ly/Adult\_uci}} from the UCI Machine Learning Repository \cite{Dua:2019}, the Compas dataset\footnote{\href{https://bit.ly/Compas_data}{https://bit.ly/Compas\_data}} introduced by ProPublica \cite{angwin2016machine}, and the HELOC dataset \cite{HELOC20XXfico}. We also employed the California Housing dataset\footnote{\href{https://bit.ly/california_statlib}{https://bit.ly/california\_statlib}} \cite{pace1997sparse} for the regression task. Supplementary details regarding each of the datasets can be found in the supplementary Section materials.

\vspace*{-0.2cm}


\paragraph{DNA datasets.} Our TT-rules framework's scalability is demonstrated using two DNA datasets, namely the single-cell RNA-seq analysis datasets for head neck, and melanoma cancer \cite{Puram2017SingleCell, Tirosh2016Dissecting} for binary classification and the TCGA lung cancer\cite{liu2018integrated} for regression. These datasets contain 23689 and 20530 features, respectively, and are commonly used in real-life machine learning applications \cite{li2017comprehensive,grewal2019application,revkov2023puree,tran2021fast}.

\vspace*{-0.2cm}

\subsection{Performances comparison - Claim A)}

\vspace*{-0.2cm}

\subsubsection{AUC/RMSE/Accuracy - Claim A-1) \& A-2)}

First, Table~\ref{tab:perfs} demonstrates that our method can handle all types of tasks, including regression, binary classification, and multi-class classification. Moreover, it outperforms most of the other interpretable methods (decision tree, RIPPER, linear/log, NAM) in various prediction tasks, except for GL~\cite{wei2019generalized}, which performs better than our method on the HELOC dataset. It is worth noting that GL does not support multi-class classification. Additionally, our method shows superior performance to more complex models such as XGBoost and DNNs on California Housing and Compas datasets. Therefore, our method can be considered comparable or superior to the current state-of-the-art methods while providing global and exact interpretability, which will be demonstrated in Section~\ref{subsec:usecase}.


\subsubsection{Complexity - Claim A-3)}

\begin{figure}[h]
\centering
\captionof{table}{Reduction of the complexity of some TT-rules models after applying optimizations from Section~\ref{subsubsec:Optimizations} on Adult~\cite{Dua:2019}, Compas~\cite{angwin2016machine} and Diabetes~\cite{Dua:2019} datasets.}
\label{tab:res_tab}
\resizebox{0.9\columnwidth}{!}{
\begin{tabular}{@{}l|cc|cc@{}}
\toprule
\textbf{Models}        &  \multicolumn{2}{c|}{TT-rules $\mathcal{R}$} & \multicolumn{2}{c}{TT-rules $\mathcal{R}_{opt}$} \\ \midrule
                    & Acc.     & Complexity                     & Acc.            & Complexity          \\ \midrule
\textbf{Adult}                & 0.846 $\pm$ 0.003   &  909 $\pm$ 212   & 0.842 $\pm$ 0.003         & 673 $\pm$ 145             \\
\textbf{Compas}    & 0.664 $\pm$ 0.013  & 343 $\pm$ 41                        & 0.664 $\pm$ 0.013  & 155 $\pm$ 22              \\
\textbf{Diabetes}    & 0.574 $\pm$ 0.008    & 22K $\pm$ 2800                         & 0.565 $\pm$ 0.009   & 15k $\pm$ 2225               \\
\bottomrule
\end{tabular}
}
\end{figure}

\paragraph{Impact of post-training optimization.} The optimizations proposed in Section~\ref{subsubsec:Optimizations} succeeded to reduce the complexity of our model as defined in Section~\ref{subsubsec:complexity} at a cost of little accuracy loss as seen in Table~\ref{tab:res_tab}. The complexity went down by a factor of $1.35\times$, $2.22\times$, and $1.47\times$ on the Adult, Compas, and Diabetes datasets respectively. The accuracy went down for the Adult and Diabetes datasets by $0.004$ and $0.009$ respectively and stayed the same for Compas.

\paragraph{Comparison with rule-based models.} Table~\ref{tab:acc_vs_complexity} presents a comparison of various rule-based models, including ours, on the Compas, Adult, and HELOC datasets, in terms of accuracy, number of rules, and complexity. We note that we report accuracy and AUC for binary classification tasks, as RIPPER and ORS do not provide probabilities. We proposed two TT-rules models: our model for high performances, as shown in Table~\ref{tab:perfs}, with floating weights, and a small model with sparse binary weights, which is also our most compact model in terms of the number of rules and complexity. Our proposed model outperforms the others in terms of accuracy on the Compas dataset and has similar performances to GL~\cite{wei2019generalized} on the Adult and HELOC datasets. Although GL provides a better tradeoff between performance and complexity, we highlight that GL does not support multi-class classification tasks and is not scalable for larger datasets such as DNA datasets, as shown in the next section. We also propose a small model as an alternative to our high-performing model. Our small model achieves accuracy that is $0.023$, $0.009$, and $0.006$ lower than our best model but requires only $3.2\times$, $2.2\times$, and $9.8\times$ fewer rules on the Compas, Adult, and HELOC datasets, respectively. We successfully reduce the complexity of our model by $14.3\times$, $34\times$, and $180\times$ on these three datasets.

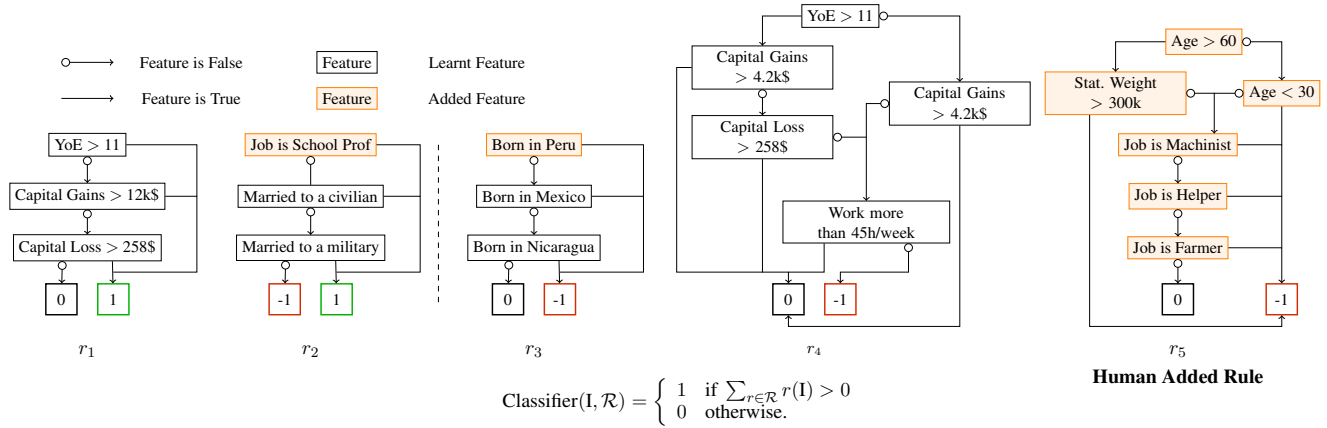
\begin{figure*}[htb!]
\vspace*{-0.2cm}
\centering
\hspace*{-0.75cm}
\resizebox{2.0\columnwidth}{!}{
\begin{tikzpicture}[every text node part/.style={align=center}]

\tikzset{
coord/.style={coordinate, on chain, on grid, node distance=6mm and 25mm},
}
    \colorlet{darkgreen}{black!30!green};
    \colorlet{darkorange}{rgb:red,5;green,1};

    \node[rectangle,draw] (L1) at (\xstart+5,1.6)  {Feature};
    \node[rectangle] (L11) at (\xstart+2.5+5,1.6) {Learnt Feature};

    \node[rectangle,draw=orange, fill=orange!10] (L2) at (\xstart+5,0.9)  {Feature};
    \node[rectangle] (L21) at (\xstart+2.5+5,0.9) {Added Feature};

    \node[coordinate] (L31) at (\xstart-0.5,1.6) {};
    \node[coordinate] (L32) at (\xstart+0.5,1.6) {};
    \draw[o->] (L31) -- (L32);
    \node[rectangle] (L33) at (\xstart+2,1.6) {Feature is False};

    \node[coordinate] (L41) at (\xstart-0.5,0.9) {};
    \node[coordinate] (L42) at (\xstart+0.5,0.9) {};
    \draw[->] (L41) -- (L42);
    \node[rectangle] (L43) at (\xstart+2,0.9) {Feature is True};

    \node[rectangle,draw] (A1) at (\xstart,0)  {YoE $>$ 11};
    \node[rectangle,draw, below of=A1] (B1) {Capital Gains $>$ 12k\$};
    \node[rectangle,draw, below of=B1] (C1) {Capital Loss $>$ 258\$};
    \node[rectangle,draw,thick,minimum height=6mm,minimum width=6mm] (D1) at (\xstart-0.5,-3)  {0};
    \node[rectangle,draw=darkgreen,thick,minimum height=6mm,minimum width=6mm] (E1) at (\xstart+0.5,-3)  {1};

    \node[rectangle] (R1) at (\xstart,-4) {\large $r_1$};

    \node[coordinate, right of=A1, node distance=60] (c11) {}; 
    \node[coordinate, above of=E1, node distance=15] (c21) {}; 
    \node[coordinate, right of=B1, node distance=60] (c31) {};

    \node[left=3.7mm of C1.south, node distance=60] (c41) {};
    \node[right=3.7mm of C1.south, node distance=60] (c51) {};

    \draw[o->] (A1.south) -- (B1);
    \draw[-] (A1.east) -| (c31) ;
    \draw[o->] (B1.south) -- (C1);
    \draw[-] (B1.east) -- (c31) |- (c21) ;
    \draw[o->] (c41.center) -- (D1);
    \draw[->] (c51.center) -- (E1);

    \node[rectangle,draw=orange, fill=orange!10] (A2) at (\xstart+4.3,0)  {Job is School Prof};
    \node[rectangle,draw, below of=A2] (B2) {Married to a civilian};
    \node[rectangle,draw, below of=B2] (C2) {Married to a military};
    \node[rectangle,draw=darkorange,thick,minimum height=6mm,minimum width=6mm] (D2) at (\xstart+4.3-0.5,-3)  {-1};
    \node[rectangle,draw=darkgreen,thick,minimum height=6mm,minimum width=6mm] (E2) at (\xstart+4.3+0.5,-3)  {1};

    \node[rectangle] (R2) at (\xstart+4.3,-4) {\large $r_2$};

    \node[coordinate, right of=A2, node distance=60] (c12) {}; 
    \node[coordinate, above of=E2, node distance=15] (c22) {}; 
    \node[coordinate, right of=B2, node distance=60] (c32) {};

    \node[left=3.7mm of C2.south, node distance=60] (c42) {};
    \node[right=3.7mm of C2.south, node distance=60] (c52) {};

    \draw[o-] (A2.south) -- (B2);
    \draw[-] (A2.east) -| (c32);
    \draw[o->] (B2.south) -- (C2);
    \draw[-] (B2.east) -- (c32) |- (c22);
    \draw[o->] (c42.center) -- (D2);
    \draw[->] (c52.center) -- (E2);

    \node[coordinate, right of=A2, node distance=70] (l1) {};
    \node[coordinate, below of=l1, node distance=90] (l2) {};

    \draw[-,dashed] (l1) -- (l2);

    \node[rectangle,draw=orange, fill=orange!10] (A3) at (\xstart+8.6,0)  {Born in Peru};
    \node[rectangle,draw, below of=A3] (B3) {Born in Mexico};
    \node[rectangle,draw, below of=B3] (C3) {Born in Nicaragua};
    \node[rectangle,draw,thick,minimum height=6mm,minimum width=6mm] (D3) at (\xstart+8.6-0.5,-3)  {0};
    \node[rectangle,draw=darkorange,thick,minimum height=6mm,minimum width=6mm] (E3) at (\xstart+8.6+.5,-3)  {-1};

    \node[rectangle] (R3) at (\xstart+8.6,-4) {\large $r_3$};

    \node[coordinate, right of=A3, node distance=60] (c13) {}; 
    \node[coordinate, above of=E3, node distance=15] (c23) {}; 
    \node[coordinate, right of=B3, node distance=60] (c33) {};

    \node[left=3.7mm of C3.south, node distance=60] (c43) {};
    \node[right=3.7mm of C3.south, node distance=60] (c53) {};

    \draw[o->] (A3.south) -- (B3);
    \draw[-] (A3.east) -| (c33);
    \draw[o->] (B3.south) -- (C3);
    \draw[-] (B3.east) -- (c33) |- (c23);
    \draw[o->] (c43.center) -- (D3);
    \draw[->] (c53.center) -- (E3);

    \node[rectangle,draw,text width=2.5cm] (A4) at (\xstart+13,1.5)  {Capital Gains $>$ 4.2k\$};
    \node[rectangle,draw, below=0.5 of A4.south,text width=2.5cm] (B4) {Capital Loss $>$ 258\$};
    \node[rectangle,draw, text width=3cm] (C4) at (\xstart+15,-1.5) {Work more than 45h/week};
    \node[rectangle,draw] (F4) at (\xstart+14.5,2.5) {YoE $>$ 11};
    \node[rectangle,draw,text width=2.5cm] (G4) at (\xstart+16.8,0.8)  {Capital Gains $>$ 4.2k\$};

    \node[rectangle] (R4) at (\xstart+14,-4) {$r_4$};
    
    \node[rectangle,draw,thick,minimum height=6mm,minimum width=6mm] (D4) at (\xstart+13+0.5,-3)  {0};
    \node[rectangle,draw=darkorange,thick,minimum height=6mm,minimum width=6mm] (E4) at (\xstart+13+1.5,-3)  {-1};

    \node[coordinate, left of=A4, node distance=47] (c14) {}; 
    \node[coordinate, above of=D4, node distance=15] (c24) {}; 
    \node[coordinate, above of=E4, node distance=15] (c34) {};
    \node[coordinate, below of=D4, node distance=15] (c64) {};
    
    \node[coordinate, right=5mm of B4.east] (c74) {};
    \node[coordinate, above of=C4] (c84) {};

    \draw [draw=none, name path=B4--c74] (B4.east) -- (c74);
    \draw [draw=none, name path=C4--c84] (c84) -- (C4.north);

    \node[coordinate, left of=c34, node distance=15] (c94) {};
    \draw [draw=none, name path=A4--B4] (A4) -- (B4);
    \draw [draw=none, name path=c34--c94] (c34) -- (c94);
    
    \node[coordinate] (inter14) at (intersection of B4--c74 and C4--c84) {};
    \node[coordinate] (inter24) at (intersection of A4--B4 and c34--c94) {};

    \node[left=7mm of C4.south, node distance=60] (c44) {};
    \node[right=7mm of C4.south, node distance=60] (c54) {};

    \draw[o->] (A4.south) -- (B4);
    \draw[-] (A4.west) -- (c14) |- (inter24) ;
    
    \draw[->] (B4.south) |- (c24) -- (D4);
    \draw[o-] (B4.east) -- (inter14);
    
    \draw[-] (c44.center) |- (c24) ;
    \draw[o->] (c54.center) |- (c34) -- (E4);
    
    \draw[o->] (F4.east) -| (G4.north);
    \draw[->] (F4.west) -| (A4.north);

    \draw[o->] (G4.west) -| (C4.north);
    \draw[->] (G4.south) |- (c64) -- (D4);

    \node[rectangle,draw=orange, fill=orange!10] (A5) at (\xstart+21,0)  {Job is Machinist};
    \node[rectangle,draw=orange, fill=orange!10, below of=A5] (B5) {Job is Helper};
    \node[rectangle,draw=orange, fill=orange!10, below of=B5] (C5) {Job is Farmer};
    \node[rectangle,draw=orange, fill=orange!10, text width=2.5cm] (F5) at (\xstart+19.8,1) {Stat. Weight $>$ 300k};
    \node[rectangle,draw=orange, fill=orange!10] (G5) at (\xstart+21.5,2) {Age $>$ 60};
    \node[rectangle,draw=orange, fill=orange!10,] (H5) at (\xstart+23,1) {Age $<$ 30};

    \node[rectangle] (R5) at (\xstart+21,-4) {\large $r_5$};
    \node[rectangle] (Human) at (\xstart+21,-4.5) {\large\textbf{Human Added Rule}};
    
    \node[rectangle,draw,thick, below of=C5,minimum height=6mm,minimum width=6mm] (D5) {0};
    \node[rectangle,draw=darkorange,thick,minimum height=6mm,minimum width=6mm] (E5) at (\xstart+23,-3)  {-1};

    \node[coordinate, right of=A5, node distance=60] (c15) {}; 
    \node[coordinate, above of=E5, node distance=15] (c25) {}; 
    \node[coordinate, right of=B5, node distance=60] (c35) {};
    \node[coordinate, right of=C5, node distance=60] (c45) {};

    \node[coordinate, below of=E5, node distance=15] (c65) {};

    \draw [draw=none, name path=c65--E5] (c65) -- (E5);
    \draw [draw=none, name path=A5--c15] (A5) -- (c15);
    \draw [draw=none, name path=B5--c35] (B5) -- (c35);
    \draw [draw=none, name path=C5--c45] (C5) -- (c45);
    
    \node[coordinate] (inter15) at (intersection of c65--E5 and A5--c15) {};
    \node[coordinate] (inter25) at (intersection of c65--E5 and B5--c35) {};
    \node[coordinate] (inter35) at (intersection of c65--E5 and C5--c45) {};

    \node[coordinate] (a51) at ($(A5.north)+(0.7,0)$) {};
    \node[coordinate] (a52) at ($(A5.north)+(0.7,1)$) {};
    \node[coordinate, right of=F5] (c75) {};
    \draw [draw=none, name path=a51--a52] (a51) -- (a52);
    \draw [draw=none, name path=F5--c75] (F5) -- (c75);
    
    \node[coordinate] (inter45) at (intersection of a51--a52 and F5--c75) {};

    \draw[o->] (A5.south) -- (B5);
    \draw[-] (A5.east) -| (inter25);
    \draw[o->] (B5.south) -- (C5);
    \draw[-] (B5.east) -| (inter35);
    \draw[o->] (C5.south) -- (D5);
    \draw[->] (C5.east) -| (E5);

    \draw[-] (H5.south) -- (inter15);
    \draw[o->] (H5.west) -| ($(A5.north)+(0.7,0)$);

    \draw[->] ($(F5.south)+(-0.5,0)$) |- (c65) -- (E5);
    \draw[o-] (F5.east) -- (inter45);

    \draw[o->] (G5.east) -| (H5);
    \draw[->] (G5.west) -| (F5);

\node[rectangle] (equ) at (\xstart+11.5, -5) {\large
$\text{Classifier}(\text{I}, \mathcal{R}) = \left\{
    \begin{array}{ll}
        1 & \mbox{if } \sum_{r \in \mathcal{R}} r(\text{I}) > 0 \\
        0 & \mbox{otherwise.}
    \end{array}
\right. 
$
};

    \node[rectangle] (empty) at (\xstart+11.5, -6) {};
    
\end{tikzpicture}}

\vspace*{-0.6cm}
\captionof{figure}{\label{fig:casestudy} Our neural network model trained on Adult dataset in the form of Boolean decision trees: the output of the DNN and the output of these decision trees are the same, reaching 83.6\% accuracy. Added Features are represented in orange rectangles. By modifying existing rules and incorporating $r_5$, the \textbf{Human Added Rule}, we reach 84.6\% accuracy. On the same test set, Random Forest reaches 85.1\% accuracy and Decision Tree 84.4\% with depth 10. There is no contradiction to the rules: one person can not be born in both Mexico and Nicaragua. The term YoE refers to the Years of Education and the Capital Gains (Losses) refer to the amount of capital gained (lost) over the year. Each rule $r_i$ is a function $r_i : \{0,1\}^n \mapsto \{-1,0,1\}$, i.e for each data sample I we associate for each rule $r_i$ a score which is in $\{-1,0,1\}$. The prediction of our classifier is then as stated above.
}
\vspace*{-0.2cm}
\end{figure*}

\subsection{Scalability - Claim B)}
\label{subsec:scalability}

Our TT-rules framework demonstrated excellent scalability to real-life datasets with up to 20K features. This result is not surprising, considering the original TTnet paper~\cite{10.1007/978-3-031-25056-9_31} showed the architecture's ability to scale to ImageNet. Furthermore, our framework's superiority was demonstrated by outperforming other rule-based models that failed to converge to such large datasets (GL~\cite{wei2019generalized}, RIPPER~\cite{cohen1995fast,cohen1999simple}). NAMs were not trained as we considered investigating the 20K graphs to be barely interpretable. Regarding performance, the TT-rules framework achieved an impressive RMSE of 0.029 on the DNA single-cell regression problem, compared to 0.092 for linear models, 0.028 for DNNs, and 0.42 for Random Forests. On the DNA multi-cross dataset, the TT-rules framework achieved an accuracy of 83.48\%, compared to 83.33\% for linear models, outperforming DNNs and Random Forests by 10.8\% and 10.4\%, respectively. Our approach not only scales but also reduces the input feature set, acting as a feature selection method. We generated a set of 1064 rules out of 20530 features for the regression problem, corresponding to a drastic reduction in complexity. For the binary classification dataset, we generated 9472 rules, which more then halved the input size from 23689 to 9472.

\vspace*{-0.2cm}

\subsection{TT-rules application case study - Claim C)}
\label{subsec:usecase}

In this section, we present the results of applying the TT-rules framework on the Adult dataset~\cite{Dua:2019}, for a specific trained example on Figure~\ref{fig:casestudy}.

\paragraph{Exact and global interpretability - Claim C-1).} For global and exact interpretability, we first apply TT-rules framework to obtain $\mathcal{R}$ and $\mathcal{R}_{opt}$. Then we transform the rules in $\mathcal{R}_{opt}$ into their equivalent ROBDD representation. This transformation is fast and automatic and  can be observed in Figure~\ref{fig:casestudy}: the resulting decision mechanism is small and easily understandable. In the Adult dataset, the goal is to predict whether an individual $I$ will earn more than \$50K per year in 1994. Given an individual's feature inputs $I$, the first rule of Figure~\ref{fig:casestudy} can be read as follows: if $I$ has completed more than 11 years of education, then the rule is satisfied. If not, then the rule is satisfied if $I$ earns more than \$4,200 in investments per month or loses more than \$228. If the rule is satisfied, $I$ earns one positive point. If $I$ has more positive points than negative points, the model predicts that $I$ will earn more than \$50K per year.

\paragraph{Human knowledge injection - Claim C-2).} Figure~\ref{fig:casestudy} illustrates our model's capability to incorporate human knowledge by allowing the modification of existing rules. However, it is important to note that we do not claim to achieve automatic human knowledge injection. The illustration simply highlights the possibility of manual rule modification in our framework.

\paragraph{Mitigating contextual drift in DNN through global and exact interpretability - Claim C-3).} It is essential to recognize that machine learning models may not always generalize well to new data from different geographic locations or contexts, a phenomenon known as ``contextual drift'' or ``concept drift''~\cite{gama2004learning}. The global and exact interpretation of DNNs is vital in this regard, as it allows for human feedback on the model's rules and the potential for these rules to be influenced by contextual drift. For example, as depicted in Figure~\ref{fig:casestudy}, this accurate model trained on US data is highly biased towards the US and is likely to perform poorly if applied in South America due to rule number 3. This highlights once again the significance of having global and exact interpretability of DNNs, as emphasized by recent NIST Artificial Intelligence Risk Management Framework~\cite{ai2023artificial}.

\vspace*{-0.2cm}

\section{Limitations and future works} \label{sec:Limitations}

Although our TT-rules framework provides a good balance between interpretability and accuracy, we observed that the generalized linear model (GL) offers a better trade-off. Specifically, for approximately the same performance, GL offers significantly less complexity. As such, future work could explore ways to identify feature interactions that work well together, similar to what GL does. Exploring automatic rule addition as an alternative to the human-based approach used in our work could also be a fruitful direction for future research.

Another interesting avenue is to apply TT-rules to time series tasks, where the interpretable rules generated by our model can provide insights into the underlying dynamics of the data. Finally, another promising area for future work would be to propose an agnostic global explainer for any model based on the TT-rules framework. 

\vspace*{-0.2cm}

\section{Conclusion} \label{sec:Conclusion}

\vspace*{-0.2cm}

In conclusion, our proposed TT-rules framework provides a new and optimized approach for achieving global and exact interpretability in regression and classification tasks. With its ability to scale on large datasets and its potential for feature reduction, the TT-rules framework appears as a valuable tool towards explainable artificial intelligence.

\bibliography{ecai}

\newpage

\appendix

\section{Datasets details}

All datasets have been split 5 times in a 80-20 train-test split for k-fold testing.

\paragraph{California Housing.} The California Housing dataset\footnote{\href{https://bit.ly/california_statlib}{https://bit.ly/california\_statlib}} is a popular dataset in machine learning and statistical modeling. It contains information about the median house value, as well as other features for census tracts in California in the early 1990s. The dataset has a total of 20,640 observations with 8 numeric input variables and 1 output variable.

\paragraph{Adult.} The Adult dataset\footnote{\href{https://bit.ly/adult_uci}{https://bit.ly/adult\_uci}} contains 48,842 individuals with 100 binary features and a label indicating whether the income is greater than 50K\$ USD or not.

\paragraph{Compas.} The Compas dataset\footnote{\href{https://bit.ly/compas_data}{https://bit.ly/compas\_data}} consists of 6,172 individuals with 17 binary features and a label that takes a value of 1 if the individual does not re-offend and 0 otherwise.

\paragraph{Heloc.} The HELOC (Home Equity Line of Credit) dataset\footnote{\href{https://community.fico.com/s/explainable-machine-learning-challenge}{https://community.fico.com/s/explainable-machine-learning-challenge}} is a collection of anonymized data from loan applications for home equity lines of credit. The dataset includes information for 10,459 recent home equity loans, with 23 numeric features and a binary label indicating

\paragraph{Diabetes.} Regarding the Diabetes dataset\footnote{\href{https://bit.ly/diabetes_130_uci}{https://bit.ly/diabetes\_130\_uci}}, it contains 100000 data points of patients with 50 features, both categorical and numerical. We kept 43 features, 5 numerical and the other are categorical which resulted in 291 binary features and 5 numerical features. We will predict one of the three labels for hospital readmission. 

\paragraph{Single-cell RNA-seq analysis.} We considered two datasets in this setting: one for head and neck cancer\footnote{\href{https://bit.ly/neck_head_rna}{https://bit.ly/neck\_head\_rna}} and another for melanoma \footnote{\href{https://bit.ly/melanoma_rna}{https://bit.ly/melanoma\_rna}} for binary classification. These datasets contains each 23689 features and a binary label. We trained and tested on the head and neck cancer dataset, then validated on the melanoma dataset. It is a classical approach in Genetics as researchers want to find general features for similar use cases.

\paragraph{TCGA lung cancer.}The TCGA lung cancer dataset\footnote{\href{https://bit.ly/tcga_lung_rna}{https://bit.ly/tcga\_lung\_rna}} is a regression dataset with 20530 features.

\section{Model architecture and training conditions}

\subsection{General}

\paragraph{Experimental environment.} The project was implemented with Python and the library PyTorch \cite{paszke2019pytorch}. Our workstation consists of eight cores Intel(R) Core(TM) i7-8650U CPU clocked at 1.90 GHz, 16 GB RAM.

\paragraph{Training method.}  We built our training method on top of \cite{jia2020efficient} project and we refer to their notations for this section. We trained the networks using the Adam optimizer \cite{kingma2014adam} for 10 epochs with a minibatch size of 128. The mean and variance statistics of batch normalization layers are recomputed on the whole training set after training finishes.

\paragraph{Weights initialization.} 
Weights for the final connected layers are initialized from a Gaussian distribution with standard deviation 0.01 and the mask weights in BinMask are enforced to be positive by taking the absolute value during initialization.

\paragraph{Other hyperparameters.} 
\label{sec:Aann_Modeldescrption3}
 We apply a weight decay of $1e-7$ on the binarized mask weight of BinMask.

\paragraph{Architecture.} Each dataset has the same input layers and output layer, with of course the size adapted to the dataset for the output layer. The first layer is a Batch Normalization layer with an $\epsilon$ of $10^{-5}$ and a momemtum of $0.1$. It is followed by the LTT block, which will be detailled for each dataset. Finally, the binary linear regression is computed, with input and output features sizes detailled below.

\subsection{Dataset Specific}

\paragraph{California Housing.} We trained the California Housing dataset in 500 epochs. The first layer is the Batch Normalization layer as described above. It is followed by a LTT block. The first convolution has 8 filters. The convolution is done with a stride of 2, a kernel size of 6 and 0 padding.

\myparagraphtitle{Adult.} We trained the Adult dataset in 10 epochs. The first layer is the Batch Normalization layer as described above. It is followed by a LTT block. The first convolution has 10 filters. The convolution is done with a stride of 5, a kernel size of 5 and no padding. It is followed by a Batch Normalization layer with the same parameters as the input one. The following convolution has 10 filters and a kernel size and a stride of 1. A last Batch Normalization finishes with the same parameter as above the LTT Block. The final linear regression takes 200 features as input and 2 as outputs. The learning rate is 0.005.

\paragraph{Compas.} We trained the Compas dataset in 60 epochs. The first layer is the Batch Normalization layer as described above. It is followed by a  LTT block. The first convolution has 20 filters (amplification of 20). The convolution is done with a stride of 1, a kernel size of 6 and no padding. It is followed by a Batch Normalization layer with the same parameters as the input one. The following convolution has 5 filters and a kernel size and a stride of 1. A last Batch Normalization finishes with the same parameter as above the LTT Block. The final linear regression takes 60 features as input and 2 as outputs. The learning rate is 0.0005. Only the best model in terms of testing accuracy was kept.

\paragraph{Heloc.} We trained the Heloc dataset in 100 epochs. The first layer is the Batch Normalization layer as described above. It is followed by a LTT block. The first convolution has 2 filters. The convolution is done with a stride of 4, a kernel size of 4 and 0 padding.

\paragraph{Diabetes.} We trained the Diabetes dataset in 10 epochs. The first layer is the Batch Normalization layer as described above. It is followed by a LTT block. The first convolution has 10 filters (amplification of 10). The convolution is done with a stride of 1, a kernel size of 6 and no padding. It is followed by a Batch Normalization layer with the same parameters as the input one. The following convolution has 10 filters and a kernel size and a stride of 1. A last Batch Normalization finishes with the same parameter as above the LTT Block. The final linear regression takes 295 features as input and 3 as outputs. The learning rate is 0.0005.

\paragraph{Single-cell RNA-seq analysis.} We trained the head and neck cancer dataset in 50 epochs. The first layer is the Batch Normalization layer as described above. It is followed by a LTT block. The first convolution has 4 filters (amplification of 4). The convolution is done with a stride of 8, a kernel size of 8 and 0 padding.

\paragraph{TCGA lung cancer.} We trained the TCGA lung cancer dataset in 100 epochs. The first layer is the Batch Normalization layer as described above. It is followed by a LTT block. The first convolution has 4 filters (amplification of 16). The convolution is done with a stride of 8, a kernel size of 9 and 0 padding.

\section{Others models training conditions}

\paragraph{Linear Models/ Decision Trees / XGBoost.} We use the sklearn implementation \cite{pedregosa2011scikit}, and tune the hyperparameters with grid search. Finally, for DT and RF, XGBoost, we tuned the minimum number of samples at leaf nodes from 1 to 100 and fixed the number of trees. 

\paragraph{NAMs.} We use the open-source implementation with the parameters specified by prior work.

\paragraph{DNNs.} We train DNNs with 10 hidden layers containing 100 units each with ReLU activation using the Adam optimizer. This architecture choice ensures that this network had the capacity to achieve perfect training accuracy on datasets used in our experiments. We use weight decay and dropout to prevent overfitting and tune hyperparameters using a similar protocol as TT-rules.

\paragraph{RIPPER / GL.} We used the implementation in \cite{arya2019one}

\end{document}